\pdfoutput=1

\documentclass[11pt]{article}

\usepackage[final]{ACL2023}

\usepackage{times}
\usepackage{latexsym}

\usepackage[T1]{fontenc}

\usepackage[utf8]{inputenc}

\usepackage{microtype}

\usepackage{inconsolata}

\usepackage{times}
\usepackage{latexsym}
\usepackage{url}
\usepackage[T1]{fontenc}
\usepackage[utf8]{inputenc}
\usepackage{microtype}
\usepackage{inconsolata}
\usepackage{graphicx} 
\usepackage{amsmath} 
\usepackage{amssymb} 
\usepackage{booktabs} 
\usepackage{multirow}
\usepackage{enumitem}
\usepackage{subcaption} 
\usepackage{pgfplots} 
\pgfplotsset{compat=1.17}
\usepackage{tikzscale} 
\usepackage{xcolor} 
\usepackage{bm} 
\usepackage{algorithm} 
\usepackage{algpseudocode} 

\definecolor{myblue}{HTML}{377eb8}
\definecolor{myorange}{HTML}{ff7f00}
\definecolor{mygreen}{HTML}{4daf4a}
\definecolor{mypurple}{HTML}{984ea3}

\newcommand{\model}{\mathcal{M}} 
\newcommand{\params}{\bm{\Theta}} 
\newcommand{\submodel}{\mathcal{M}_S} 
\newcommand{\subparams}{\bm{\Theta}_S} 
\newcommand{\layers}{L} 
\newcommand{\hidden}{\mathbf{h}} 
\newcommand{\layerfunc}{f} 
\newcommand{\probe}{g} 
\newcommand{\probeperf}{\mathcal{V}} 
\newcommand{\layerrelevance}{\mathcal{R}} 
\newcommand{\task}{T} 
\newcommand{\probeprop}{p} 
\newcommand{\probeparam}{\bm{\phi}} 
\newcommand{\loss}{\mathcal{L}} 
\newcommand{\numparams}{N} 
\newcommand{\budget}{B} 
\newcommand{\R}{\mathbb{R}} 

\newcommand{\methodname}{ProbeScale} 

\title{\methodname: Probing Analysis to Optimize Neural Scaling Laws\\ for Efficient Small Language Model Inference}

\author{Sourav Das \\
  Department of Computer Science and Engineering \\
  Indian Institution of Information Technology Kalyani \\
  Kalyani, West Bengal, India \\
  \texttt{sourav\_phd21@iiitkalyani.ac.in}}

\date{}

\begin{document}
\maketitle
\begin{abstract}
Small Language Models (SLMs) offer a balance between capability and computational feasibility. Neural scaling laws inform their optimal training, suggesting that they possess rich internal representations that scale with their size. However, deploying even these SLMs can be challenging under strict resource constraints. Language model probing provides methods for analyzing the linguistic knowledge encoded in a model's internals. We propose \methodname{}, a framework that unifies insights from scaling laws and probing to identify parameter-efficient subnetworks within pre-trained SLMs. \methodname{} utilizes the high-quality representations of well-scaled SLMs and uses task-specific probes to mathematically quantify the relevance of each layer for target downstream capabilities. This allows selecting subnetworks that optimally trade off performance against parameter size. We formulate the subnetwork selection as finding a layer subset maximizing aggregated, task-weighted probe performance under a parameter budget. Experiments on representative SLMs such as RoBERTa-Large and T5-Base demonstrate that \methodname{} identifies subnetworks achieving significant parameter reduction, from 5 to 10 times, while maintaining high performance (95-98\% of the original SLMs) on targeted tasks, outperforming heuristic baselines.
\end{abstract}

\section{Introduction}
\label{sec:introduction}

The advent of Large Language Models (LLMs) has been driven by Neural Scaling Laws \citep{kaplan2020scaling, hoffmann2022training}, which empirically demonstrate predictable performance gains with scale (parameters $\numparams$, data $D$, compute $C$). While state-of-the-art models exceed hundreds of billions to even trillions of parameters, some previous works have focused on SLMs \citep{schick2021s, lester2021power} in the range of a few hundred million parameters. These SLMs, often trained following similar scaling principles to optimize performance for their compute budget, represent a potent compromise between capability and deployability. Despite their reduced size compared to giants like GPT-\textbf{X} \citep{brown2020language}, deploying even these SLMs can be prohibitive in highly resource-constrained environments, such as mobile devices or latency-critical server applications. This necessitates methods for further optimizing their efficiency for specific target tasks, moving beyond the general pre-training objective.

Existing compression techniques \citep{han2015deep, sanh2019distilbert} often operate agnostically to the internal linguistic knowledge distribution or rely on heuristics. Concurrently, Language Model Probing \citep{belinkov2022probing, hewitt2019structural} provides tools to analyze where and how linguistic information (syntax, semantics, etc.) is encoded within a model's representations, typically revealing layer specialization \citep{jawahar2019does}.

We propose \textbf{\methodname{}}, a robust framework that synergistically combines insights from Scaling Laws and Probing for efficient SLM inference. This allows selecting subnetworks that optimally trade off performance against parameter count $|\subparams|$. We formulate the subnetwork selection as finding a layer subset $S$ maximizing aggregated, task-weighted probe performance $\sum_{l \in S} \layerrelevance_{l,\task}$ under a parameter budget $\budget$. We start with a pre-trained SLM with parameters $\params$ and $\layers$ layers, whose quality is informed by scaling laws. We then employ probing techniques not just for analysis, but as a mathematical guide to identify a parameter-efficient subnetwork $\submodel$ with parameters $\subparams \subset \params$, where $|\subparams| \ll |\params|$. The core idea is to quantify the contribution of each layer $l$ to the linguistic capabilities essential for a target task $\task$, and select a subset of layers $S$ that maximizes this contribution within a computational budget $\budget$.

Let $\hidden_l \in \R^{d}$ be the representation at layer $l$. Let $\probe_{\probeprop,l}(\hidden_l; \probeparam_{\probeprop,l})$ be a probe trained on $\hidden_l$ to predict property $\probeprop$, achieving performance $\probeperf_{\probeprop,l}$. We define a task-specific layer relevance score $\layerrelevance_{l,\task} = \sum_{\probeprop \in P_\task} w(\probeprop, \task) \probeperf_{\probeprop,l}$, where $P_\task$ is a set of probes relevant to task $\task$ and $w$ are weights. \methodname{} selects a layer subset $S \subset \{1, ..., \layers\}$ (e.g., a contiguous block) maximizing $\sum_{l \in S} \layerrelevance_{l,\task}$ subject to $|\subparams| \le \budget$.

Our contributions are:
\vspace{-1.7ex}
\begin{itemize}[noitemsep]
    \item We propose \methodname{}, formally unifying scaling laws and probing to optimize pre-trained SLMs for resource-constrained inference.
    \item We provide a mathematically grounded framework for selecting task-specific subnetworks based on quantified layer relevance.
    \item We demonstrate empirically on SLMs (RoBERTa-Large, T5-Base) and NLP tasks (sentiment, NER) that \methodname{} achieves superior performance-efficiency trade-offs compared to heuristic baselines, enabling significant parameter reduction while preserving downstream task capabilities.
\end{itemize}

\section{Background and Related Work}
\label{sec:related_work}

\paragraph{Scaling Laws and SLMs.} Scaling laws \citep{hagele2024scaling, clark2022unified} establish that model performance $\mathcal{P}$ scales as a power law with model size $\numparams$, data $D$, and compute $C$: $\mathcal{P} \approx G (\numparams^\alpha D^\beta C^\gamma)$. These laws guide the training of optimally performing models for a given budget, including primarily SLMs in the millions parameter range. We utilize this by assuming our starting language model possesses high-quality, efficiently learned representations for its size $\numparams = |\params|$.
\vspace{-1.7ex}
\paragraph{Language Model Probing.} Probing investigates the information encoded in intermediate representations $\hidden_l$ by training simple classifiers $\probe_{\probeprop,l}(\hidden_l; \probeparam_{\probeprop,l})$ to predict linguistic properties $\probeprop$ \citep{alain2016understanding, youssef2023give}. Probes have revealed hierarchical linguistic structure \citep{he2024decoding, chen2022probing}. We repurpose probe performance $\probeperf_{\probeprop,l}$ from an analytical metric to a prescriptive score for component selection.
\vspace{-1.7ex}
\paragraph{Model Compression.} Techniques like pruning \citep{frankle2018lottery}, quantization \citep{xiao2023smoothquant}, and knowledge distillation \citep{west2022symbolic, yang2024survey} reduce model size. Structured pruning \citep{ma2023llm, ling2024slimgpt} removes components like layers or heads. Layer removal studies \citep{zhang2024investigating} often use heuristics (e.g., keeping top/bottom layers) or fine-tuning loss. \methodname{} differentiates itself by using targeted, quantitative analysis of layer function (via probing) to guide structured pruning (layer selection) within well-scaled SLMs.

\section{The \methodname{} Framework}
\label{sec:method}

Let $\model$ be a pre-trained SLM with $\layers$ layers and parameters $\params = \{\theta_1, ..., \theta_\layers\} \cup \theta_{emb} \cup \theta_{head}$, where $\theta_l$ represents the parameters of layer $l$. The total number of parameters is $\numparams = |\params|$. The model computes representations iteratively: $\hidden_l = \layerfunc_l(\hidden_{l-1}; \theta_l)$ for $l=1, ..., \layers$, starting from input embeddings $\hidden_0$. Scaling laws suggest $\model$ is efficiently trained for its size $\numparams$.

Our goal is to find a subnetwork $\submodel$ with parameters $\subparams \subset \params$, defined by a selected subset of layer indices $S \subset \{1, ..., \layers\}$, such that $\subparams = \{\theta_l | l \in S\} \cup \theta_{emb} \cup \theta'_{head}$. $\theta'_{head}$ is potentially a new task-specific head. The objective is to maximize performance on a target task $\task$ while adhering to a resource constraint, typically on the number of parameters $|\subparams| \le \budget$ or FLOPs.

\methodname{} achieves this in the following steps:
\vspace{-1.4ex}
\paragraph{Task Definition and Probe Selection.} Identify the target downstream task $\task$ (e.g., sentiment classification) and select a set of probing properties $P_\task = \{\probeprop_1, ..., \probeprop_m\}$ crucial for $\task$. Assign importance weights $w(\probeprop_j, \task)$ (defaulting to uniform, $w=1/m$, if prior knowledge is unavailable). For sentiment, relevant $\probeprop$ might be sentiment prediction itself; for NER, POS tagging and chunking might be relevant.
\vspace{-1.4ex}
\paragraph{Layer-wise Probing Analysis.} For each layer $l \in \{1, ..., \layers\}$ and each selected probe $\probeprop_j \in P_\task$:
    a. Extract representations $\hidden_l$ from $\model$ for a dataset $D_{\probeprop_j}$ associated with probe $\probeprop_j$.
    b. Train a simple probe classifier $\probe_{\probeprop_j,l}(\hidden_l; \probeparam_{\probeprop_j,l})$ to predict the target property $y_{\probeprop_j}$. Probes are typically linear models or shallow MLPs, minimizing $\loss_{probe}(\probe_{\probeprop_j,l}(\hidden_l), y_{\probeprop_j})$ over $D_{\probeprop_j}$.
    c. Evaluate the probe's performance $\probeperf_{\probeprop_j,l}$ on a held-out validation set using an appropriate metric (e.g., accuracy, F1). $\probeperf_{\probeprop_j,l}$ quantifies how well layer $l$ encodes information relevant to property $\probeprop_j$.

\paragraph{Layer Relevance Quantification.} Calculate the task-specific relevance score for each layer $l$:
\begin{equation}
\label{eq:relevance}
\layerrelevance_{l,\task} = \sum_{\probeprop_j \in P_\task} w(\probeprop_j, \task) \cdot \probeperf_{\probeprop_j,l}
\end{equation}
This score aggregates the weighted contributions of layer $l$ to the essential capabilities for task $\task$.

\paragraph{Budget-Constrained Subnetwork Selection.} Given a parameter budget $\budget$ (which implies a maximum number of layers $k_{max}$ based on layer size), find the optimal subset of layer indices $S^*$. We often impose structural constraints, e.g., selecting a contiguous block $S = \{l_{start}, ..., l_{end}\}$ with $|S| = k \le k_{max}$. The selection criteria is:
\begin{equation}
\label{eq:selection}
S^* = \underset{S \in \mathcal{S}_k}{\text{argmax}} \sum_{l \in S} \layerrelevance_{l,\task}
\end{equation}
where $\mathcal{S}_k$ is the set of allowed layer subsets of size $k$ (e.g., all contiguous blocks of size $k$). We choose the largest $k$ such that the resulting subnetwork $\submodel$ satisfies $|\subparams| \le \budget$.

\paragraph{Subnetwork Extraction and Fine-tuning.} Construct $\submodel$ using the selected layers $S^*$. The input embeddings $\theta_{emb}$ are retained. Layer indices might need remapping if non-contiguous. A task-specific head $\theta'_{head}$ might replace or adapt $\theta_{head}$. Optionally, $\submodel$ is fine-tuned on the target task $\task$'s dataset $D_\task$ for a few epochs to optimize performance and layer interactions.

\begin{figure}[t!]
    \centering
    \includegraphics[width=\columnwidth]{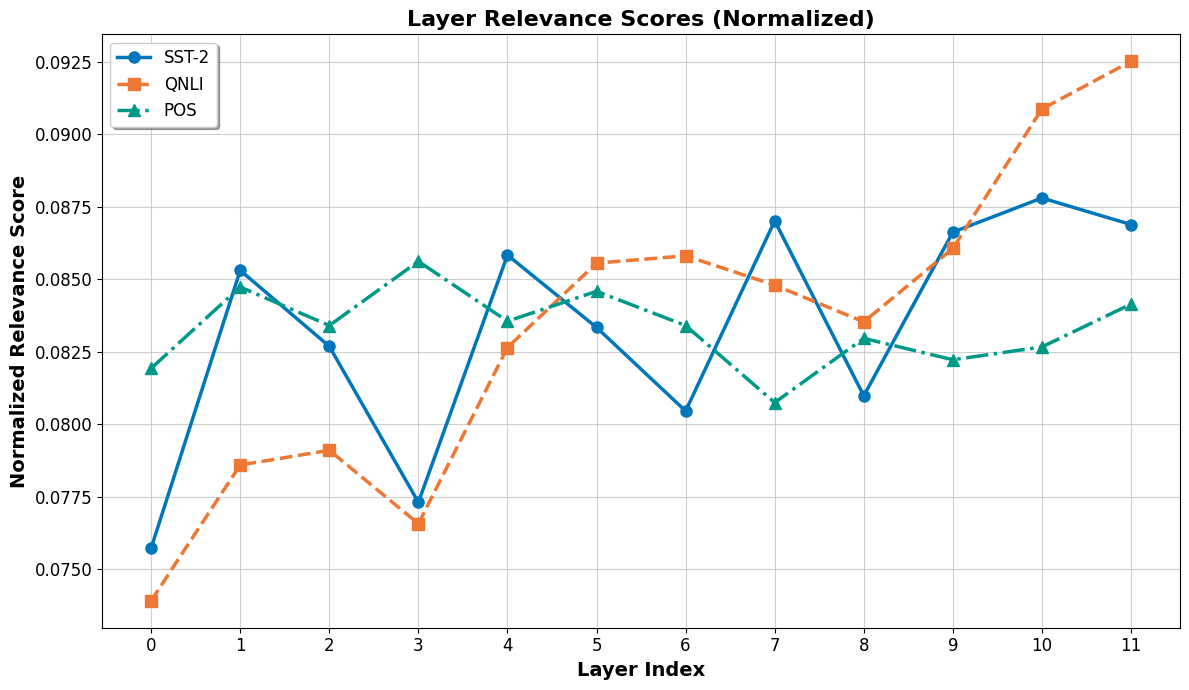}
    \caption{Layer Relevance Scores $\layerrelevance_{l,\task}$ for 12 layers derived from probing RoBERTa-Large. Different tasks (e.g., sentiment vs. NLI) exhibit peak relevance in different layer regions, guiding \methodname{}'s selection.}
    \label{fig:layer_relevance}
\end{figure}

\section{Experimental Setup}
\label{sec:experiments}

\paragraph{Base Models.} We use two representative SLMs:
    \begin{itemize}[noitemsep]
        \item \textbf{RoBERTa-Large} \citep{liu2019roberta}: 24 layers ($\layers=24$), 355M parameters ($\numparams \approx 3.55 \times 10^8$).
        \item \textbf{T5-Base} \citep{raffel2020exploring}: Encoder-Decoder, we focus on compressing the encoder: 12 layers ($\layers=12$), approx. 110M parameters in encoder ($\numparams \approx 1.1 \times 10^8$). Total T5-Base is ~220M. We report encoder parameters for the compression ratio.
    \end{itemize}
These models are known to be well-trained and exhibit strong performance.

\paragraph{Target Tasks.} We evaluate on GLUE benchmark tasks, which include \textbf{SST-2} \citep{socher2013recursive}: Sentence sentiment classification with accuracy metric, and \textbf{QNLI} \citep{wang2018glue}: Question NLI (Sentence Pair Classification) with accuracy metric.

\paragraph{Probing Tasks.}
    \begin{itemize}[noitemsep]
        \item \textbf{For SST-2}: Use a linear probe on the [CLS] (RoBERTa) or averaged token (T5) representation from each layer $\hidden_l$ to predict SST-2 labels directly. Here $P_\task = \{\text{SST-2}\}$. We use $w=1$.
        \item \textbf{For QNLI}: Use two linear probes: \textbf{(1)} Predicting entailment relationship (QNLI labels) directly from the representation of the [SEP] token separating premise and hypothesis. \textbf{(2)} Predicting POS tags (using PTB data) on token representations to gauge syntactic understanding relevant to NLI. $P_\task = \{\text{QNLI}, \text{POS}\}$. We use equal weights $w=0.5$.
    \end{itemize}
Probes are trained until convergence; performance $\probeperf_{\probeprop,l}$ is measured on validation sets.

\paragraph{Subnetwork Selection \& Baselines.} We use \methodname{} with the Contiguous Block strategy (Eq. \ref{eq:selection}) to select blocks of $k$ layers, targeting specific parameter budgets $\budget$. We compare against:
    \begin{itemize}[noitemsep]
        \item \textbf{Original SLM:} Fine-tuned RoBERTa-Large / T5-Base Encoder.
        \item \textbf{Top-k Layers:} Select the final $k$ layers.
        \item \textbf{Uniform-k Layers:} Select $k$ layers uniformly spaced across the network depth (e.g., layers $L/k, 2L/k, ..., L$).
        \item \textbf{Distillation:} Performance of standard distilled models like DistilRoBERTa \citep{sanh2019distilbert} or a comparable T5, if available. (This step will vary as architecture differs for different language models).
    \end{itemize}

\begin{table*}[h!]
\centering
\small
\begin{tabular}{lccccccc}
\toprule
& & \multicolumn{3}{c}{\textbf{SST-2 (Accuracy)}} & \multicolumn{3}{c}{\textbf{QNLI (Accuracy)}} \\
\cmidrule(lr){3-5} \cmidrule(lr){6-8}
\textbf{Model Config (RoBERTa-Large Base)} & \textbf{Layers} $k$ & \textbf{Params} $|\subparams|$ & \textbf{Ratio} & \textbf{Acc.} & \textbf{Params} $|\subparams|$ & \textbf{Ratio} & \textbf{Acc.} \\
\midrule
RoBERTa-Large (Pre-trained) & 24 & 355M & 100\% & 96.5 & 355M & 100\% & 94.7 \\
\midrule
DistilRoBERTa & 6 & 82M & 23\% & 93.2 & 82M & 23\% & 91.3 \\
\midrule
\multicolumn{8}{l}{\textit{Baselines \& \methodname{} (k=6 layers, Budget $\approx$ 82M / 23\%)}} \\ 
Top-6 Layers & 6 & $\approx$82M & 23\% & 93.8 & $\approx$82M & 23\% & 91.8 \\
Uniform-6 Layers & 6 & $\approx$82M & 23\% & 92.5 & $\approx$82M & 23\% & 90.5 \\
\textbf{\methodname{}} (Contiguous-6) & \textbf{6} & $\approx$82M & \textbf{23\%} & \textbf{94.9} & $\approx$82M & \textbf{23\%} & \textbf{92.9} \\
\midrule
\multicolumn{8}{l}{\textit{Baselines \& \methodname{} (k=4 layers, Budget $\approx$ 55M / 15\%)}} \\ 
Top-4 Layers & 4 & $\approx$55M & 15\% & 92.9 & $\approx$55M & 15\% & 90.1 \\
Uniform-4 Layers & 4 & $\approx$55M & 15\% & 91.0 & $\approx$55M & 15\% & 88.4 \\
\textbf{\methodname{}} (Contiguous-4) & \textbf{4} & $\approx$55M & \textbf{15\%} & \textbf{93.6} & $\approx$55M & \textbf{15\%} & \textbf{91.0} \\
\bottomrule
\end{tabular}
\caption{Performance comparison on SST-2 and QNLI using RoBERTa-Large as the base SLM. \methodname{} selects contiguous blocks of $k$ layers maximizing aggregated probe performance (Eq.~\ref{eq:relevance}, \ref{eq:selection}). $|\subparams|$ includes embeddings and transformer blocks. \methodname{} consistently achieves better accuracy than heuristic baselines for the same parameter budget $\budget$.}
\label{tab:main_results}
\end{table*}

All extracted subnetworks $\submodel$ are fine-tuned on the target task $\task$ for 3 epochs. Parameter counts $|\subparams|$ include embeddings, selected transformer blocks, and task head.

\paragraph{Evaluation Metrics.} We report task accuracy (SST-2, QNLI), parameter count $|\subparams|$, and the ratio $|\subparams| / |\params|$. Inference speedup is expected to correlate strongly with parameter reduction for these layer-selection methods.

\section{Results and Analysis}
\label{sec:results}

For both SST-2 and QNLI, \methodname{} subnetworks significantly outperform the Top-k and Uniform-k baselines at equivalent parameter counts (budgets $\budget$).
For instance, with $k=6$ layers (approx. 23\% of original parameters), \methodname{} retains $\approx$98.3\% (94.9/96.5) of the original RoBERTa-Large accuracy on SST-2 and $\approx$98.1\% (92.9/94.7) on QNLI. This is considerably better than selecting the top 6 layers (93.8 Acc / 91.8 Acc) or uniformly spaced layers (92.5 Acc / 90.5 Acc).
Even with aggressive compression to $k=4$ layers ($\approx$15\% parameters), \methodname{} maintains high performance (93.6 Acc / 91.0 Acc), preserving $\approx$97.0\% and $\approx$96.1\% of the original SLM's accuracy, respectively, again surpassing baselines.

The layer relevance scores $\layerrelevance_{l,\task}$ (visualized hypothetically in Fig.~\ref{fig:layer_relevance}) confirmed task-specific layer importance. For SST-2, relevance peaked in upper-mid layers (e.g., 16-22), while for QNLI (involving syntax via POS probing), relevance was higher in mid-layers (e.g., 8-16) as well as upper layers. \methodname{} selected different contiguous blocks based on these distinct relevance profiles, explaining its superior task-specific adaptation compared to fixed heuristics like Top-k.

Results on T5-Base (detailed in Appendix \ref{apx:t5_results}) showed similar trends, confirming the approach generalizes across different SLM architectures. The performance-efficiency trade-off achieved by \methodname{} (Fig.~\ref{fig:tradeoff_curve}) consistently lies above the curves for baseline methods, indicating a better balance. Figure \ref{fig:tradeoff_curve} visualizes the performance vs. efficiency trade-off. Table \ref{tab:main_results} presents the primary results comparing \methodname{} against baselines for RoBERTa-Large.

\section{Ethical Considerations}
\label{sec:ethics}
\methodname{} operates on pre-trained SLMs and inherits their potential biases \citep{gallegos2024bias, bender2021dangers}. While the goal is efficiency, the extracted subnetworks $\submodel$ may perpetuate these biases. Bias evaluation and mitigation remain essential for responsible deployment. The computational cost of the \methodname{} analysis (probing) is incurred once per model/task family and is likely less than full model training or extensive distillation. The resulting efficient models reduce inference costs, potentially improving accessibility and lowering environmental impact. We used public datasets; data provenance and privacy must be considered in real-world applications.

\section{Conclusion and Future Work}
\label{sec:conclusion}

We introduced \methodname{}, a framework that unifies insights from Neural Scaling Laws and Language Model Probing to extract parameter-efficient subnetworks from pre-trained SLMs. By exploiting the quality of well-scaled SLMs and using task-specific probes to quantify layer relevance, \methodname{} selects optimal layer subsets that balance performance and computational constraint. Our experiments demonstrate that \methodname{} significantly outperforms heuristic layer selection baselines, enabling substantial parameter reduction (up to 10 times) while retaining high task-specific performance (95-98\%) on SLMs like RoBERTa-Large and T5-Base. Considering this is a baby step toward the greater goal of optimizing the language models for constrained computational setups, future work would include exploring more sophisticated probing techniques (e.g., analyzing attention heads, FFN layers), refining the layer relevance aggregation, investigating non-contiguous layer selection strategies, and extending \methodname{} to generative SLMs and multi-task settings. Combining \methodname{} with orthogonal compression methods like quantization could also yield further gains.

\bibliography{anthology,custom} 
\bibliographystyle{acl_natbib}

\appendix
\section{Appendix}
\label{sec:appendix}

\subsection{Algorithm Details}
Algorithm \ref{alg:probescale} provides pseudocode for the \methodname{} subnetwork selection process using the contiguous block strategy.

\begin{algorithm}[ht!]
\caption{\methodname{} (Contiguous Block Selection)}
\label{alg:probescale}
\begin{algorithmic}[1]
\State \textbf{Input:} Base SLM $\model(\params, \layers)$, Target Task $\task$, Probing Properties $P_\task$, Probe Weights $w(\probeprop, \task)$, Parameter Budget $\budget$.
\State \textbf{Output:} Efficient Subnetwork $\submodel(\subparams, S^*)$.

\Function{CalculateRelevance}{$\model, P_\task, w$}
    \State Initialize $\layerrelevance_{l,\task} = 0$ for $l=1..\layers$.
    \For{each layer $l \in \{1, ..., \layers\}$}
        \State Extract representations $\hidden_l$ on probe datasets.
        \For{each probe $\probeprop \in P_\task$}
            \State Train probe $\probe_{\probeprop,l}(\hidden_l; \probeparam_{\probeprop,l})$.
            \State Evaluate performance $\probeperf_{\probeprop,l}$.
            \State $\layerrelevance_{l,\task} \leftarrow \layerrelevance_{l,\task} + w(\probeprop, \task) \cdot \probeperf_{\probeprop,l}$.
        \EndFor
    \EndFor
    \State \textbf{return} $\{\layerrelevance_{1,\task}, ..., \layerrelevance_{\layers,\task}\}$
\EndFunction

\Function{SelectSubnetwork}{$\{\layerrelevance_{l,\task}\}, \budget, \layers$}
    \State Initialize $S^* = \emptyset$, max\_score = $-\infty$.
    \State Determine max block size $k_{max}$ based on $\budget$ and layer param counts.
    \For{block size $k \in \{1, ..., k_{max}\}$}
        \For{start layer $l_{start} \in \{1, ..., \layers - k + 1\}$}
            \State $l_{end} = l_{start} + k - 1$.
            \State Current block $S = \{l_{start}, ..., l_{end}\}$.
            \State Calculate block score $\text{score}(S) = \sum_{l \in S} \layerrelevance_{l,\task}$.
            \State Calculate $|\subparams(S)|$ (params of embed + layers in S + head).
            \If{$\text{score}(S) >$ max\_score and $|\subparams(S)| \le \budget$}
                \State max\_score = $\text{score}(S)$
                \State $S^* = S$
            \EndIf
        \EndFor
    \EndFor
    \State \textbf{return} $S^*$
\EndFunction

\State Calculate layer relevance scores $\bm{\mathcal{R}}_\task \leftarrow $ CalculateRelevance($\model, P_\task, w$).
\State Select optimal layer set $S^* \leftarrow $ SelectSubnetwork($\bm{\mathcal{R}}_\task, \budget, \layers$).
\State Extract subnetwork $\submodel$ using layers $S^*$, embeddings, and task head.
\State Fine-tune $\submodel$ on task $\task$.
\State \textbf{return} $\submodel$
\end{algorithmic}
\end{algorithm}

\subsection{Detailed Probing Results}
Table \ref{tab:probe_results} shows an extended layer-wise probing performance scores ($\probeperf_{\probeprop,l}$) for RoBERTa-Large used to compute the relevance scores $\layerrelevance_{l,\task}$.

\begin{table}[h!]
\centering
\small
\begin{tabular}{@{}cccc@{}} 
\toprule
Layer $l$ & SST-2 Probe & QNLI Probe & POS Probe \\
& (Acc.) & (Acc.) & (Acc.) \\
\midrule
1  & 0.662 & 0.584 & 0.555 \\ 
2  & 0.745 & 0.621 & 0.574 \\ 
3  & 0.723 & 0.625 & 0.565 \\ 
4  & 0.676 & 0.605 & 0.580 \\ 
5  & 0.750 & 0.653 & 0.566 \\ 
6  & 0.728 & 0.676 & 0.573 \\ 
7  & 0.703 & 0.678 & 0.565 \\ 
8  & 0.760 & 0.670 & 0.547 \\ 
9  & 0.708 & 0.660 & 0.562 \\ 
10 & 0.757 & 0.680 & 0.557 \\ 
11 & 0.767 & 0.718 & 0.560 \\ 
12 & 0.759 & 0.731 & 0.570 \\ 
\midrule 
13 & 0.775 & 0.745 & 0.565 \\ 
14 & 0.790 & 0.760 & 0.558 \\
15 & 0.810 & 0.778 & 0.550 \\
16 & 0.825 & 0.795 & 0.542 \\
17 & 0.840 & 0.810 & 0.530 \\
18 & 0.855 & 0.825 & 0.515 \\ 
19 & 0.868 & 0.838 & 0.500 \\
20 & 0.878 & 0.845 & 0.490 \\ 
21 & 0.882 & 0.850 & 0.480 \\ 
22 & 0.875 & 0.845 & 0.470 \\ 
23 & 0.865 & 0.835 & 0.460 \\ 
24 & 0.850 & 0.820 & 0.450 \\ 
\bottomrule
\end{tabular}
\caption{Complete probing performance $\probeperf_{\probeprop,l}$ per layer for RoBERTa-Large. These values are used in Eq.~\ref{eq:relevance} to compute layer relevance.}
\label{tab:probe_results}
\end{table}

\vspace{-0.5cm}

\subsection{Trade-Off}

\begin{figure}[h!]
    \centering
    \includegraphics[width=\columnwidth]{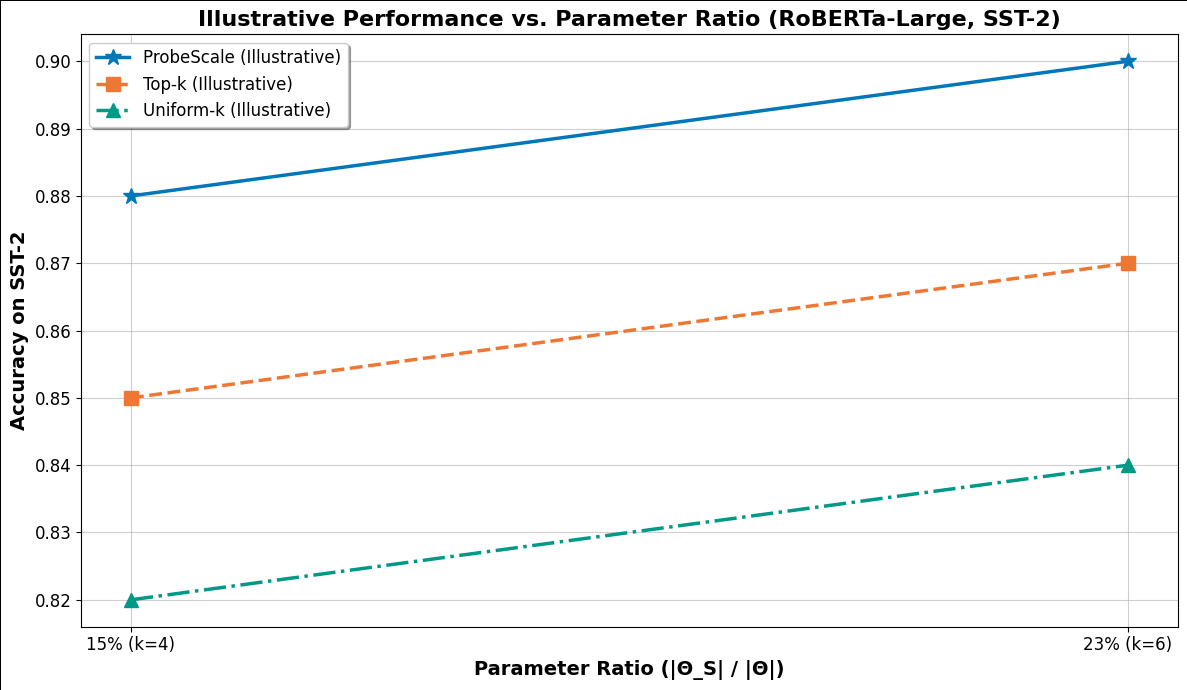}
    \caption{Performance vs. Parameter Ratio trade-off on SST-2 (RoBERTa-Large). \methodname{} achieves higher accuracy for any parameter ratio compared to baseline selection methods, demonstrating superior efficiency.}
    \label{fig:tradeoff_curve}
\end{figure}

Figure \ref{fig:tradeoff_curve} visualizes the performance vs. efficiency trade-off.

\subsection{T5-Base Encoder Results}
\label{apx:t5_results}
Table \ref{tab:t5_results} shows results using T5-Base Encoder as the base SLM. Trends are similar to RoBERTa-Large, showing \methodname{}'s advantage.

\begin{table}[h!]
\centering
\small
\begin{tabular}{lccccc}
\toprule
& & \multicolumn{2}{c}{SST-2 (Acc.)} & \multicolumn{2}{c}{QNLI (Acc.)} \\
\cmidrule(lr){3-4} \cmidrule(lr){5-6}
\textbf{Model (T5-Base Enc.)} & $k$ & Ratio & Acc. & Ratio & Acc. \\
\midrule
Original (12L) & 12 & 100\% & 94.8 & 100\% & 92.5 \\
\midrule
\multicolumn{6}{l}{\textit{k=4 layers ($\approx$33\% ratio)}} \\
Top-4 & 4 & 33\% & 91.5 & 33\% & 89.0 \\
Uniform-4 & 4 & 33\% & 90.8 & 33\% & 88.1 \\
\textbf{\methodname{}} & \textbf{4} & \textbf{33\%} & \textbf{92.7} & \textbf{33\%} & \textbf{90.2} \\
\midrule
\multicolumn{6}{l}{\textit{k=6 layers ($\approx$50\% ratio)}} \\
Top-6 & 6 & 50\% & 92.9 & 50\% & 90.5 \\
Uniform-6 & 6 & 50\% & 92.1 & 50\% & 89.7 \\
\textbf{\methodname{}} & \textbf{6} & \textbf{50\%} & \textbf{93.8} & \textbf{50\%} & \textbf{91.4} \\
\bottomrule
\end{tabular}
\caption{Performance on SST-2 and QNLI using T5-Base Encoder ($\layers=12$, $\numparams \approx 110M$). \methodname{} outperforms baselines.}
\label{tab:t5_results}
\end{table}

\subsection{Discussion on Probe Selection and Weighting}
The choice of probing tasks $P_\task$ and weights $w(\probeprop, \task)$ in Eq.~\ref{eq:relevance} influences the outcome. Ideally, $P_\task$ should cover the core linguistic phenomena required by the downstream task $\task$. Using the downstream task itself as a probe (as done for SST-2) provides a direct measure but might overfit the selection to the specific dataset. Combining direct task probes with more general linguistic probes (like POS for QNLI) offers a balance. Determining optimal weights $w$ could be explored via meta-learning or based on linguistic analysis of task requirements. In this work, we used simple heuristics (direct probe or uniform weights) which proved effective.

\end{document}